\documentclass[letterpaper]{article}
\usepackage{times}
\usepackage{helvet}
\usepackage{courier}
\usepackage{amssymb}
\usepackage{graphicx}
\usepackage{pstricks,pst-node,pst-tree}
\usepackage{enumerate} 
\newtheorem{theorem}{Theorem}
\newtheorem{definition}{Definition}

\begin{document}

\title{Mixed Integer Linear Programming For Exact Finite-Horizon Planning In Decentralized Pomdps}
\author{Raghav Aras \and Alain Dutech \and Fran\c{c}ois Charpillet\\
INRIA-Lorraine / Loria,\\
615 rue du Jardin Botanique,\\
54602 Villers-l\`es-Nancy,
France \\
$\{$aras, dutech, charp$\}$@loria.fr
}
\maketitle
\begin{abstract}
\begin{quote}
We consider the problem of finding an $n$-agent joint-policy for the optimal finite-horizon control of a decentralized Pomdp (Dec-Pomdp).  This is a problem of very high complexity (NEXP-hard in $n \geq 2$).  In this paper,  we propose a new mathematical programming approach for the problem.  Our approach is based on two ideas:  First,  we represent each agent's policy in the sequence-form and not in the tree-form,  thereby obtaining a very compact representation of the set of joint-policies.  Second,  using this compact representation,  we solve this problem as an instance of combinatorial optimization for which we formulate a mixed integer linear program (MILP).  The optimal solution of the MILP directly yields an optimal joint-policy for the Dec-Pomdp.  Computational experience shows that formulating and solving the MILP requires significantly less time to solve benchmark Dec-Pomdp problems than existing algorithms.  For example, the multi-agent tiger problem for horizon $4$ is solved in 72 secs with the MILP whereas existing algorithms require several hours to solve it.   
\end{quote}
\end{abstract}

\section{Introduction}
In a \textit{finite-horizon} Dec-Pomdp \cite{2002Bernstein},  a set of $n$ agents cooperate to control a Markov decision process for $\kappa$ steps under two constraints:  \textit{partial observability} and \textit{decentralization}.  Partial observability signifies that the agents are imperfectly informed about the state of the process during control.  Decentralization signifies that the agents are \textit{differently} imperfectly informed during the control.  The agents begin the control of the process with the same,  possibly imperfect,  information about the state.  During the control each agent receives \textit{private} information about the state of the process, which he cannot divulge to the other agents.  The agents' private information can have an impact on what they collectively do.  Thus,  \textit{before} the control begins,  each agent must reason not only about the possible states of the process during the control (as in a Pomdp) but he must also reason about the information that \textit{could} be held by other agents during the control.  In effect,  the agent must also reason about which policies the other agents would use.  Partial observability and decentralization make Dec-Pomdps very difficult to solve.  Finding an optimal solution to a Dec-Pomdp is NEXP-hard in the number of agents \cite{2002Bernstein};  finding a locally optimal solution to a Dec-Pomdp is NP-hard in the size of the Dec-Pomdp problem (determined by $\kappa$ and the sizes of the sets of joint-actions and joint-observations) \cite{1992Koller}. 
\subsection{Motivation for a new approach}
The three existing exact algorithms DP \cite{2004Hansen},  MAA$^{*}$ \cite{2005Szer} and PBDP \cite{2006Szer} are able to solve only very small Dec-Pomdps in reasonable time ($2$ agents, horizon $\leq 4$,  action and observation set sizes $\leq 3$).  Their lack of scalability is predictable from the negative complexity results.    Therefore,  the question is not so much whether these algorithms can be improved upon in the absolute,  but rather if a relative improvement can be achieved.  In other words,  can we push the computational envelop a bit further on this problem?  In this paper,  we present a new approach based on \textit{integer programming},   which does manifest a much superior performance in practice than the existing algorithms.  For instance,  through our approach,  the multi-agent Tiger problem \cite{2003Nair} for horizon $4$ can be solved in 72 seconds as against the few hours required by the PBDP algorithm \cite{2006Szer} (the only current algorithm able to solve this instance).  Similarly,  the MABC problem \cite{2004Hansen} for horizon $5$ is solved in 25 seconds as against the $10^{5}$ seconds required by PBDP. So we might tentatively answer in the positive to the above question.  There is of course a more relevant reason for pushing this envelop.  The three algorithms serve as a basis for approximate algorithms such as Approximate-DP \cite{2004Montemerlo} and MBDP \cite{2007Seuken},  and these seem to scale to much longer horizons and to much larger problems.  So,  a more efficient exact algorithm is important from this perspective as well. We discuss this in more detail in the last section.   
\subsection{A new,  mixed integer programming approach}
Existing Dec-Pomdp algorithms represent an agent's policy as a \textit{tree} and a joint-policy as a tuple of policy-trees.  The size of the set of policy-trees of each agent is \textit{doubly exponential} in the horizon.  Hence,  the set of joint-policies is doubly exponential in the horizon and exponential in the number of agents.  This adversely impacts the space and time requirements of the algorithms.  In our approach we discard the tree representation in favor of the \textit{sequence-form} representation which was introduced in a seminal paper on computational game theory \cite{1994Koller}.  In the sequence-form,  every finite-horizon \textit{deterministic} policy of an agent can be represented as a \textit{subset} of the set sequences of actions and observations of the agent.  The problem of finding an optimal \textit{deterministic} joint-policy is thus equivalent to the problem of finding for each agent a subset from a larger set.  This problem thus becomes an instance of \textit{combinatorial optimization} and we conceive a mixed integer linear program (MILP) to solve it.  The key insight of Koller's approach (and therefore of our approach) is that the size of the set of sequences from each subset is drawn is only exponential in the horizon and not doubly exponential in it,  as is the case with the size of the set of policy trees.  This allows us to formulate an MILP whose size is exponential in $\kappa$ and $n$.  For small problems such as MA-Tiger and MABC, it is feasible to represent the MILP in memory.  Furthermore,  and equally importantly,  the constraints matrix of the MILP is \textit{sparse}.  The consequence of this is that in practice the MILP is solved very quickly (in the order of seconds).  Thus,  we have an effective method to compute an optimal deterministic finite-horizon joint-policy.  Restricting attention to deterministic joint-policies does not limit the applicability of our approach in any way since in every finite-horizon Dec-Pomdp there exists at least one optimal joint-policy that is deterministic.   It is also not evident that relaxing this restriction has any benefit.  Implicitly,  existing algorithms also restrict attention to deterministic joint-policies.  In this paper `policy' and `joint-policy' shall mean deterministic policy and deterministic joint-policy respectively unless otherwise specified.

\section{The finite-horizon Dec-Pomdp problem}
A finite-horizon Dec-Pomdp problem is defined by the following elements.  We are given $N$,  a set of $n$ agents and $S$, a set of states.  The $n$ agents in $N$ are numbered from $1$ to $n$.  The states are numbered from $1$ to $|S|$.  For each $i$th agent,  we are given $A_{i}$,  the agent's set of actions and $\Omega_{i}$,  his set of observations.  The cross-product $A_{1}$ $\times$ $A_{2}$ $\ldots$ $\times$ $A_{n}$ is called the set of \textit{joint-actions} and it is denoted by $A$.  Similarly,  the cross-product $\Omega_{1}$ $\times$ $\Omega_{2}$ $\ldots$ $\times$ $\Omega_{n}$ is called the set of \textit{joint-observations} and it is denoted by $\Omega$.  The joint-actions are numbered from $1$ to $|A|$ and the joint-observations are numbered from $1$ to $|\Omega|$.  Then,  we are given for each $a$th joint-action, the matrices $T^{a}$,  $Z^{a}$ and the vector $R^{a}$:
\begin{enumerate}[(a)]
\item  $T^{a}_{ss'}$ is the probability of transitioning to the $s'$th state if the agents take the $a$th joint-action in $s$th state.  
\item  $Z^{a}_{s'o}$ is the probability of the agents receiving the $o$th joint-observation and transitioning to $s'$th if they take the $a$th.
\item  $R^{a}_{s}$ is the real-valued reward the agents obtain if they take the $a$th joint-action in the $s$th state.
\end{enumerate}
We are given $b_{0}$, which represents the initial \textit{belief state} and it is common knowledge amongst the agents.  A belief state is a probability distribution over $S$.  In a belief state $b$,  the probability of the $s$th state is denoted by $b[s]$.   Finally,  we are given $\kappa \geq 1$, a finite number that is the \textit{horizon} of the control.  The control of the Dec-Pomdp is described as follows.  At each step $t$ of $\kappa$ steps:  the agents take a joint-action,  they receive a joint-observation,  they receive a common reward $r_{t}$,  and the process transitions to a new belief state as a function of the previous belief state,  the joint-action and the joint-observation.  However,  at each step,  agents do not reveal to one another the actions they take and observations they receive at that step or at previous steps.  Since an agent does not know the actions taken by the other agents and the observations received by the other agents during the $\kappa$ steps,  at each step he takes actions strictly as a function of the actions he has taken previously and observations he has received previously.  This function is called his \textit{policy}.  To control the Dec-Pomdp for $\kappa$ steps,  each agent requires a $\kappa$-step policy, henceforth written as $\kappa$-policy.  The tuple of the agents' policies forms a \textit{joint-policy}.  An optimal joint-policy is one which maximizes $E(\sum_{t = 1}^{\kappa}r_{t})$,  the sum of expected rewards the agents obtain for the $\kappa$ steps.
\subsection{Policy in the tree-form}
\begin{figure}
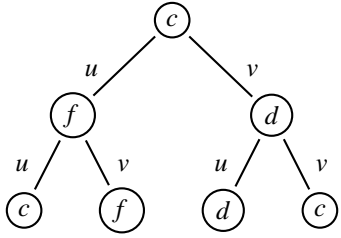

\begin{center}
\pstree[nodesep=2pt,levelsep=8ex]{\Tcircle{\textit{c}} }{
        \pstree{ \Tcircle{\textit{f}}\tlput{\textit{u}} }{
	             \Tcircle{\textit{c}}\tlput{\textit{u}}
                     \Tcircle{\textit{f}}\trput{\textit{v}}
                              }
	\pstree{ \Tcircle{\textit{d}}\trput{\textit{v}} }{
	             \Tcircle{\textit{d}}\tlput{\textit{u}}
                     \Tcircle{\textit{c}}\trput{\textit{v}}
                              }
                    }

\end{center}
\caption{A $3$-step policy $\psi$.}
\label{FigurePolicy}
\end{figure}

The canonical representation of a policy, used in existing Dec-Pomdp algorithms,  is the \textit{tree-form}.  In this form,  a $\kappa$-policy of the $i$th agent can be represented as a rooted tree with $\kappa$ levels in which each non-terminal node has $|\Omega_{i}|$ children.  This tree is called a $\kappa$-policy-tree.  Each node is labeled by an action to take and each edge is labeled by an observation that may occur.  Using a policy-tree,  during the control,  the agent follows a path from the root to a leaf depending on the observations he receives.  An example of a policy-tree is shown in Figure \ref{FigurePolicy}.  The number of nodes in a $\kappa$-policy-tree of the $i$th agent is $\frac{|\Omega_{i}|^{\kappa} - 1}{|\Omega_{i}| - 1}$.  It is thus exponential in $\kappa$.  For example,  with $|\Omega_{i}| = 2$,  a $3$-policy-tree,  as the one shown in Figure \ref{FigurePolicy}, has $\frac{2^{3} - 1}{2 - 1} = 7$ nodes.  The set of $\kappa$-policy-trees of the $i$th agent is the set of all the $\frac{|\Omega_{i}|^{\kappa} - 1}{|\Omega_{i}| - 1}$ sized permutations of the actions in $A_{i}$.  Therefore,  the size of the set of $\kappa$-policy-trees of the $i$th agent is $|A_{i}|^\frac{|\Omega_{i}|^{\kappa} - 1}{|\Omega_{i}| - 1}$,  doubly exponential in $\kappa$.   
\section{Policy in the sequence-form}
The double exponentiality associated with the set of policy-trees can be avoided by using the \textit{sequence-form} representation of a policy.  We begin a description of this representation by defining a sequence.
\begin{definition}
A {\rm{sequence of length $t$ of the $i$th agent}} is an ordered list of 2$t$ - 1 elements, $t \geq 1$, in which the elements in odd positions are actions from $A_{i}$ and those in even positions are observations from $\Omega_{i}$.  
\end{definition}
Thus,  in a sequence of length $t$ there are $t$ actions and $t$ - 1 observations.  The shortest possible sequence is of length $1$,  which consists of just an action and no observations.  We denote the set of all possible sequences of length $t$ which can be conceived from $A_{i}$ and $O_{i}$ by ${\cal{S}}^{t}_{i}$.  We denote the set ${\cal{S}}^{1}_{i}$ $\cup$ ${\cal{S}}^{2}_{i}$ $\cup$ $\ldots$ ${\cal{S}}^{\kappa}_{i}$ by ${\cal{S}}_{i}$.  We shall now see how a $\kappa$-policy can be represented as a set of sequences,  or more precisely as a subset of ${\cal{S}}_{i}$.  Assume that $\kappa = 3$ and the policy-tree $\psi$ shown in Figure \ref{FigurePolicy} is of the $i$th agent.  Starting from the root-node and descending down the edges of the tree,  we can enumerate the sequences of this tree.  The first sequence we obtain is in the root-node itself, the sequence consisting of the action $c$ and no observations.  This is a sequence of length 1.  Then,  going down the edge labeled by $u$ from the root-node,  we come to the node labeled by the action $f$.  At this point, we obtain a second sequence $cuf$, which is of length 2.  It has two actions and one observation.  Similarly,  taking the other edge from the root-node,  we come to the node labeled by $d$ and obtain a third sequence $cvd$, also of length 2.  When all the leaves of the tree have been visited,  the set of sequences we obtain is,
\begin{eqnarray*}
{\cal{S}}(\psi) = \mbox{$\Big\{$ $c$, $cuf$, $cvd$, $cufuc$, $cufvf$, $cvdud$, $cvdvc$ $\Big\}$}
\end{eqnarray*}
This set contains $1$ sequence of length $1$,  $2$ sequences of length $2$ and $4$ sequences of length $3$ to give a total of $7$ sequences corresponding to the $7$ nodes in $\psi$.  It is evident that the set ${\cal{S}}(\psi)$ is equivalent to the policy-tree $\psi$.  That is,  \textit{given} the set ${\cal{S}}(\psi)$,  the agent can use it as a $3$-step policy.  As this simple exercise shows,  any finite-step policy can be written as a finite set of sequences.  Now,  ${\cal{S}}(\psi)$ \textit{is a subset of} ${\cal{S}}_{i}$,  the set of all possible sequences of lengths less than or equal to $3$,  and so is every $3$-policy of the $i$th agent.  Thus, for any given value of $\kappa$,  every $\kappa$-policy of the $i$th agent is a subset of ${\cal{S}}_{i}$.  This is main idea of the sequence-form representation of a policy.     
\subsection{Policy as a vector}
We can streamline the subset-set relationship between a $\kappa$-policy and ${\cal{S}}_{i}$ by representing the former as a vector of binary values. Let the sequences in ${\cal{S}}_{i}$ be numbered from $1$ to $|{\cal{S}}_{i}|$.  Since every $\kappa$-policy of the $i$th agent is a subset of ${\cal{S}}_{i}$,  every sequence in ${\cal{S}}_{i}$ is either in the policy or it is not.  Thus a $\kappa$-policy of the $i$th agent can be represented as a $|{\cal{S}}_{i}|$-vector of binary values $0$ or $1$,  such that if the $j$th sequence in ${\cal{S}}_{i}$ is in the policy then the $j$th element of the vector equals $1$ and if it is not,  then the $j$th element of the vector equals $0$.  Let the set of $|{\cal{S}}_{i}|$-vectors of binary values $0$ or $1$ be denoted by ${\cal{X}}_{i}$.  Thus every $\kappa$-policy of the $i$th agent is member of the set ${\cal{X}}_{i}$.  Let $p$ be the $j$th sequence in ${\cal{S}}_{i}$.  For a vector $x_{i} \in {\cal{X}}_{i}$,  value of the $j$th element in $x_{i}$ shall be conveniently represented as $x_{i}[p]$.

\subsection{Policy constraints of the $i$th agent}
\label{SubsectionPolicyConstraints}
Thus,  every $\kappa$-policy of the $i$th agent is a member of ${\cal{X}}_{i}$.  The inverse of this is course untrue;  not every member of ${\cal{X}}_{i}$ is a $\kappa$-policy.  We therefore need to define which vectors in ${\cal{X}}_{i}$ can represent a $\kappa$-policy.  We shall give a more general definition,  one that includes stochastic policies as well as deterministic policies.  We shall in fact define which vectors in  $\mathbb{R}^{|{\cal{S}}_{i}|}$ represent a $\kappa$-step policy, be it a stochastic policy or a deterministic one.  The definition takes the form of a system of linear equations which must be satisfied by a vector in $\mathbb{R}^{|{\cal{S}}_{i}|}$ if it is to represent a $\kappa$-policy.  Given a sequence $p$, an action $a$ and an observation $o$,  let $poa$ denote the sequence obtained on appending $o$ and $a$ to the end of $p$.  Let ${\cal{S}}_{i}^{'}$ denote the set ${\cal{S}}^{1}_{i}$ $\cup$ ${\cal{S}}^{2}_{i}$ $\cup$ $\ldots$ ${\cal{S}}^{\kappa - 1}_{i}$.
\begin{definition}
Let $|{\cal{S}}_{i}| = z$.  A vector $w \in \mathbb{R}^{z}$ is a $\kappa$-step, possibly stochastic, policy of the $i$th agent if,
\begin{eqnarray}
\label{PolicyConstraints1}
&&\sum_{a \in A_{i}}w[a] = 1\\
\label{PolicyConstraints2}
&&w[p] - \sum_{a \in A_{i}}w[poa] = 0,  \mbox{\quad $\forall$ $p \in {\cal{S}}_{i}^{'}$, $o \in \Omega_{i}$}\\
\label{PolicyConstraints3}
&&w \geq 0
\end{eqnarray}
\end{definition}
We call the system of linear equations (\ref{PolicyConstraints1})-({\ref{PolicyConstraints3}) the \textit{policy constraints of the $i$th agent}.  Policy constraints recreate the tree structure of a policy.   They appear in a slightly different form, as Lemma 5.1 in \cite{1994Koller}.  We can write the policy constraints in the matrix form as $C_{i}w = b_{i}$, $w \geq 0$, where $C_{i}$ is the matrix of the coefficients of the variables in the equations (\ref{PolicyConstraints1})-({\ref{PolicyConstraints2}) and $b_{i}$ is a vector of appropriate length whose first element is $1$ and the remaining elements are $0$,  representing the r.h.s of the equations.  Note that it is implicit in the above definition that the value of each element of $w$ is constrained to be in the interval [0,1].  Hence,  we can define a deterministic $\kappa$-policy of the $i$th agent as follows.
\begin{definition}
A vector $x_{i} \in {\cal{X}}_{i}$ is a $\kappa$-policy of the $i$th agent if $C_{i}x_{i} = b_{i}$.
\end{definition}
We shall call a policy represented as a vector as a \textit{policy-vector} just to distinguish it from a policy-tree.  The representation of a policy as a policy-vector is in fact the sequence-form representation we have been alluding to.  Given a vector from $x_{i} \in {\cal{X}}_{i}$ which satisfies the policy constraints,  the agent can use it just as he would use as a policy-tree without requiring any additional book-keeping.  Let choosing a sequence mean taking the last action in the sequence.  In using $x_{i}$,  at the first step,  he chooses the action $a$ such that $x_{i}[a] = 1$.  There will be only one such action.  Then on receiving an observation, say $o$,  he chooses the sequence $aoa'$ such that $x_{i}[aoa'] = 1$.  Again there will be only one such sequence.  In general,  if at step $t$ he has chosen the sequence $p$ and then received the observation $o$,  then he chooses the unique sequence $poa''$ such that $x_{i}[poa''] = 1$ at the $(t + 1)$th step.  Thus,  at each step,  the agent must know the sequence of actions he has taken and the sequence of observations he has received till that step in order to know which action to take according to $x_{i}$.  This requirement is called \textit{perfect recall} in game theory,  and it is implicit in the use of a policy-tree.

\subsection{Advantage of the sequence-form representation}
The size of ${\cal{S}}^{t}_{i}$ is $|A_{i}|^{t}|\Omega_{i}|^{t - 1}$.    The size of ${\cal{S}}_{i}$ is thus $\sum_{t = 1}^{\kappa}|A_{i}|^{t}|\Omega_{i}|^{t - 1}$,  exponential in $\kappa$.  Since every $\kappa$-policy is in theory available if the set ${\cal{S}}_{i}$ is available,  the latter serves as a \textit{search space} for $\kappa$-policies of the $i$th agent.  The good news is of course that this search space is only exponential in $\kappa$.  This compares favorably with the search space represented by the set of $\kappa$-policy-trees which is doubly exponential in $\kappa$.  We thus have at our disposal an exponentially smaller space in which to search for an agent's policy.  More precisely,  to find a $\kappa$-policy of the $i$th agent,  we need to set up and solve the system the policy constraints.  The number of equations in this system is $c_{i}$ $=$ $1$ + $\sum_{t = 1}^{\kappa - 1}|A_{i}|^{t}|\Omega_{i}|^{t}$.  $C_{i}$ is thus a $c_{i} \times |{\cal{S}}_{i}|$ matrix.  Now notice that $C_{i}$ is a \textit{sparse} matrix,  that is,  it has only a very small number of nonzero entries per row or column,  while most of its entries are $0$s.  In $C_{i}$,  the number of nonzero entries per row is only $1 + |A_{i}|$, and it is constant per row.  Sparse matrices are typically easier to solve that dense matrices of the same size.  The relatively small size of $C_{i}$ and its sparsity combine to form a relatively efficient method to find a $\kappa$-policy of the $i$th agent.

\section{Value of a Joint-policy}
The agents control the the finite-horizon Dec-Pomdp by a $\kappa$-step joint-policy, henceforth written as a $\kappa$-joint-policy.  A joint-policy is just the tuple formed by the agents' individual policies. Thus,  a $\kappa$-joint-policy is an $n$-tuple of $\kappa$-policies.  A $\kappa$-joint-policy may be an $n$-tuple of $\kappa$-policy-trees or it may be an $n$-tuple of $\kappa$-policy-vectors.  Given a joint-policy $\pi$ in either representation,  the policy of the $i$th agent in it shall be denoted by $\pi_{i}$.  A joint-policy is evaluated by computing its \textit{value}.  The value of a joint-policy represents the sum of expected rewards the agents obtain if it is executed starting from the given initial belief state $b_{0}$.  The value of a joint-policy $\pi$ shall be denoted by ${\cal{V}}(\pi)$.     

\subsection{Value of a joint-policy as an $n$-tuple of policy-trees}
Given a $t$-policy $\sigma$ of an agent,  $t \leq \kappa$, let $a(\sigma)$ denote the action in the root node of $\sigma$ and let $\sigma(o')$ denote the sub-tree attached to the root-node of $\sigma$ into which the edge labeled by the observation $o'$ enters.  Furthermore,  given a $t$-joint-policy $\pi$, let $a(\pi)$ denote the joint-action ($a(\pi_{1})$, $a(\pi_{2})$, $\ldots$, $a(\pi_{n})$) and given a joint-observation $o$,  let $\pi(o)$ denote the ($t-1$)-joint-policy ($\pi_{1}(o_{1})$, $\pi_{2}(o_{2})$, $\ldots$, $\pi_{n}(o_{n})$).  Now let $\pi$ be a $\kappa$-joint-policy which is an $n$-tuple of $\kappa$-policy trees.  The value of $\pi$ is expressed in terms of the $\kappa$-step \textit{value-function} of the Dec-Pomdp denoted by $V^{\kappa}$ as follows,   
\begin{equation}
\label{EquationPolicyTreeValue}
{\cal{V}}(\pi) = \sum_{s \in S}b_{0}[s]V^{\kappa}(s, \pi)
\end{equation} 
in which $V^{\kappa}$ is expressed recursively as,
\begin{equation}
V^{\kappa}(s, \pi) = R^{a(\pi)}_{s} + \sum_{o \in \Omega}\sum_{s' \in S}T^{a(\pi)}_{ss'}Z^{a(\pi)}_{s'o}V^{\kappa - 1}(s', \pi(o))
\end{equation}
For $t = 1$,  $V^{t}(s, a) = R^{a}_{s}$.  An \textit{optimal} $\kappa$-joint-policy is one whose value is the maximum.

\subsection{Value of a joint-policy as an $n$-tuple of policy-vectors}
The value of a $\kappa$-joint-policy that is an $n$-tuple of policy-vectors is expressed in terms of the values of its \textit{joint-sequences}.  A joint-sequence is defined analogously to a sequence.
\begin{definition}  
A {\rm{joint-sequence of length $t$}} is an ordered list of 2$t$ - 1 elements, $t \geq 1$, in which the elements in odd positions are joint-actions from $A$ and those in even positions are joint-observations from $\Omega$.  
\end{definition}
Equivalently,  we can also define a joint-sequence of length $t$ as an $n$-tuple of sequences of length $t$.  Given a joint-sequence $q$,  the sequence of the $i$th agent in $q$ shall be denoted by $q_{i}$.  The set of joint-sequences of length $t$,  denoted by ${\cal{S}}^{t}$,  is thus the cross-product set ${\cal{S}}_{1}^{t}$ $\times$ ${\cal{S}}_{2}^{t}$ $\ldots$ $\times$ ${\cal{S}}_{n}^{t}$.  Given a joint-sequence $q$ of length $t$,  the $(j \leq t)$th joint-action in it shall be denoted by $a^{j}_{q}$ and the $(h < t)$th joint-observation in it shall be denoted by $o^{h}_{q}$.  We now define the value of a joint-sequence.
\subsection{Joint-sequence value}
The value of a joint-sequence $q$ of length $t$,  denoted by $\nu(q)$, is \textit{independent} of any joint-policy.  It is simply a property of the Dec-Pomdp model.  It is a product of two quantities:  $\rho(q)$,  the probability of $q$ occurring and ${\cal{R}}(q)$,  the sum of expected rewards the joint-actions in $q$ obtain:
\begin{equation}
\nu(q) = \rho(q){\cal{R}}(q)
\end{equation}
These quantities are defined and computed as follows.  $\rho(q)$ is the probability,
\begin{eqnarray}
\rho(q) &=& \mbox{Pr}(o^{1}_{q}, o^{2}_{q}, \ldots, o^{t - 1}_{q}|b_{0}, a^{1}_{q}, a^{2}_{q}, \ldots, a^{t - 1}_{q})\\
&=& \prod_{j = 1}^{t - 1}\mbox{Pr}(o^{j}_{q}|b_{0}, a^{1}_{q}, o^{1}_{q}, \ldots, o^{j - 2}_{q}, a^{j - 1}_{q})\\ 
&=& \prod_{j = 1}^{t - 1}\mbox{Pr}(o^{j}_{q}|b_{j - 1}^{q})
\end{eqnarray}
where $b_{j - 1}^{q}$ is a belief state which, if computed as follows, serves a \textit{sufficient statistic} for the joint-sequence ($a^{1}_{q}$, $o^{1}_{q}$, $\ldots$, $o^{j - 2}_{q}$, $a^{j - 1}_{q}$).  Let $o$ denote $o^{j}_{q}$ and $a$ denote $a^{j}_{q}$.  Let $b_{j - 1}^{q}$ be given.  Then,  
\begin{equation}
\label{EquationProbablityObservation}
\mbox{Pr}(o|b_{j - 1}^{q}) = \sum_{s \in S}\sum_{s' \in S} b_{j - 1}^{q}[s] T^{a}_{ss'}Z^{a}_{s'o}\\
\end{equation}
and $b_{j}^{q}$ is given as,  (for each $s \in S$),
\begin{equation}
\label{EquationBeliefStateUpdate}
b_{j}^{q}[s] = \frac{\sum_{s' \in S} b_{j - 1}[s'] T^{a}_{s's}Z^{a}_{so}}{\mbox{Pr}(o|b_{j - 1}^{q})}
\end{equation}
Thus $\rho(q)$ is computed as follows.  We assign $b_{0}$ to $b_{0}^{q}$.  For each non-zero $j < t$,  we calculate Pr($o^{j}_{q}|b_{j - 1}^{q}$) using eq. (\ref{EquationProbablityObservation}).  If for any $t$,  we find that Pr($o^{j}_{q}|b_{j - 1}^{q}$) is $0$, we set $\rho(q)$ to $0$ and terminate.  On the other hand,  whenever Pr($o^{j}_{q}|b_{j - 1}^{q}$) $>$ $0$,  we compute $b_{j}^{q}[s]$ for each state $s \in S$ using eq.(\ref{EquationBeliefStateUpdate}) and continue.  The quantity ${\cal{R}}(q)$ is simply the sum of the expected rewards the joint-actions in $q$ obtain in the belief states $b_{j}^{q}$s.  Assigning,  as before,  $b_{0}$ to $b_{0}^{q}$, and denoting $a^{j}_{q}$ by $a$,
\begin{eqnarray}
{\cal{R}}(q) = \sum_{j = 1}^{t}\sum_{s \in S}b_{j - 1}^{q}[s]R^{a}_{s}
\end{eqnarray}
Recall that fundamentally,  a $\kappa$-policy is just a set of sequences of different lengths.  Given a policy $\sigma$ of the $i$th agent let the subset of $\sigma$ containing sequences of length $t$ be denoted by $\sigma^{t}$.  Then given a joint-policy $\pi$,  the set of joint-sequences of length $t$ of $\pi$ is simply the set $\pi_{1}^{t}$ $\times$ $\pi_{2}^{t}$ $\times$ $\ldots$ $\times$ $\pi_{n}^{t}$.  Note that if a joint-sequence $q$ is in $\pi$,  then $\prod_{i = 1}^{n}\pi_{i}[q_{i}] = 1$ and if it is not,  then $\prod_{i = 1}^{n}\pi_{i}[q_{i}] = 0$.  We can now define the value of a $\kappa$-joint-policy $\pi$ in terms of the values of the joint-sequences.  In particular, we need consider only joint-sequences of length $\kappa$.  Thus,    
\begin{equation}
\label{EquationJointPolicyValue}
{\cal{V}}(\pi) = \sum_{q \in {\cal{S}}^{\kappa}}\nu(q)\prod_{i = 1}^{n}\pi_{i}[q_{i}]
\end{equation} 
The derivation of eq. (\ref{EquationJointPolicyValue}) from eq. (\ref{EquationPolicyTreeValue}) is quite straightforward and is omitted.

\section{Algorithm}
We shall now describe a mixed integer linear program (MILP) that finds an optimal $\kappa$-joint-policy.  We start our description with the following naive mathematical program (MP) which just implements the definition of an optimal $\kappa$-joint-policy.  This implies finding for each agent $i$ a vector $\overline{x}_{i} \in \mathbb{R}^{|{\cal{S}}_{i}|}$ which satisfies the policy constraints of the $i$th agent and the quantity ${\cal{V}}$($\overline{x}_{1}$, $\overline{x}_{2}$, $\ldots$, $\overline{x}_{n}$) is maximized.  Letting $\overline{x}$ $=$ ($\overline{x}_{1}$, $\overline{x}_{2}$, $\ldots$, $\overline{x}_{n}$), the naive MP, denoted by $\mathtt{MP}$-$\mathtt{Dec}$,  is as follows,
\begin{eqnarray}
\label{NLP1}
\mbox{maximize $\quad$} f(\overline{x}) &\equiv& \sum_{q \in {\cal{S}}^{\kappa}}\nu(q)\prod_{i = 1}^{n}\overline{x}_{i}[q_{i}]\\
\mbox{s.t. $\quad$ $\forall$ $i \in N$:}&& C_{i}\overline{x}_{i} = b_{i}\\ 
\label{NLP2}
&& \overline{x}_{i} \geq 0
\end{eqnarray}
An optimal solution to $\mathtt{MP}$-$\mathtt{Dec}$ would yield an optimal (possibly,  stochastic) $\kappa$-joint-policy $\overline{x}$.  However,  since $f(\overline{x})$ is a nonconcave, nonlinear function,  not only is solving $\mathtt{MP}$-$\mathtt{Dec}$ NP-hard,  but more importantly,  it is also not possible to guarantee finding a globally optimal $\kappa$-joint-policy.  A simple fix to get rid of the nonconcave nonlinear $f(\overline{x})$ in  $\mathtt{MP}$-$\mathtt{Dec}$ is to somehow linearize $f(\overline{x})$,  that is,  to transform it into a linear function.  Linearization of $f(\overline{x})$ is achieved by using more  variables and more constraints in addition to those in $\mathtt{MP}$-$\mathtt{Dec}$.    The additional variables pertain to joint-sequences and the additional constraints are required to relate the variables of joint-sequences to those of sequences.  The linearization of $f(\overline{x})$ takes place in three steps.  At the end of the three steps,  $\mathtt{MP}$-$\mathtt{Dec}$ is converted to an integer linear program (ILP) on which the proposed MILP is based.

\subsection{Linearization of $f(\overline{x})$:  step $1$}
The simple idea in linearizing a nonlinear function is to use a variable for each nonlinear term that appears in the function.  In the case of $f(\overline{x})$,  the nonlinear terms are, for each joint-sequence $q$ of length $\kappa$, $\prod_{i = 1}^{n}\overline{x}_{i}[q_{i}]$.  Therefore,  to replace the nonlinear terms in $f(\overline{x})$,  we need to use a variable for every joint-sequence $q$ of length $\kappa$.  Let $\overline{y}[q] \geq 0$ be the variable for $q$ and let, 
\begin{eqnarray}
\label{EquationJustY}
f(\overline{y}) \equiv \sum_{q \in {\cal{S}}^{\kappa}}\nu(q)\overline{y}[q]
\end{eqnarray}
So the first step in linearizing $f(\overline{x})$ is to change the objective in $\mathtt{MP}$-$\mathtt{Dec}$ to $f(\overline{y})$ and introduce the $|{\cal{S}}^{\kappa}|$-vector $\overline{y} \geq 0$ of variables in it.  We denote this modified MP by $\mathtt{MP1}$-$\mathtt{Dec}$.    
\subsection{Linearization of $f(\overline{x})$:  step $2$}
Once the objective is changed to $f(\overline{y})$,  we need to relate the variables representing joint-sequences ($\overline{y}$) to those representing agents' sequences (the $\overline{x}_{i}$ vectors).  In other words,  we need to add the following constraints to $\mathtt{MP1}$-$\mathtt{Dec}$,
\begin{equation}
\label{EquationXToY}
\prod_{i = 1}^{n}\overline{x}_{i}[q_{i}] = \overline{y}[q], \mbox{$\quad$ $\forall$ $q \in {\cal{S}}^{\kappa}$}
\end{equation} 
But the constraints (\ref{EquationXToY}) are \textit{nonconvex}.  So,  if they are added to $\mathtt{MP1}$-$\mathtt{Dec}$,  it would amount to maximizing a linear function under nonconvex, nonlinear constraints,  and again we would not have any guarantee of finding the globally optimal solution.  We therefore must also linearize these constraints.  We shall do this in this step and the next.  Suppose that ($\overline{x}_{1}$, $\overline{x}_{2}$, $\ldots$, $\overline{x}_{n}$) is a solution to $\mathtt{MP1}$-$\mathtt{Dec}$.  Then,  for each joint-sequence $q$ of length $\kappa$,  $\prod_{i = 1}^{n}\overline{x}_{i}[q_{i}]$ takes a value in [0,1].  In other words,  it can take an infinite number of values.  We can limit the values it can take by requiring that the vectors $\overline{x}_{i}$ be vectors of \textit{binary} variables, $0$ or $1$.  Moreover,  since we want $\prod_{i =1}^{n}\overline{x}_{i}[q_{i}]$ to equal $\overline{y}[q]$, but want to avoid the constraints (\ref{EquationXToY}),  we should also require that each $\overline{y}$ variable be a binary variable.  Thus,  the second step in linearizing $f(\overline{x})$ is to add the following constraints to $\mathtt{MP1}$-$\mathtt{Dec}$:
\begin{eqnarray}
\label{EquationIntegerConstraints1}
\overline{x}_{i}[p] \in \{0, 1\}, &&\mbox{$\forall$ $i \in N$, $\forall$ $p \in {\cal{S}}_{i}$}\\
\label{EquationIntegerConstraints2}
\overline{y}[q] \in \{0, 1\}, &&\mbox{$\forall$ $q \in {\cal{S}}^{\kappa}$}
\end{eqnarray} 
Note that with these constraints in $\mathtt{MP1}$-$\mathtt{Dec}$, $\overline{x}_{i}$ would represent a deterministic $\kappa$-policy of the $i$th agent.  Constraints (\ref{EquationIntegerConstraints1})-(\ref{EquationIntegerConstraints2}) are called \textit{integer constraints}.  We denote the MP formed by adding integer constraints to $\mathtt{MP1}$-$\mathtt{Dec}$ by $\mathtt{MP2}$-$\mathtt{Dec}$.

\subsection{Linearization of $f(\overline{x})$:  step $3$}
This is key step in the linearization.  The number of sequences of length $\kappa$ in a $\kappa$-policy of the $i$th agent is $\tau_{i} = |\Omega_{i}|^{\kappa - 1}$.  Hence the number of joint-sequences of length $\kappa$ in a $\kappa$-joint-policy $\tau = \prod_{i = 1}^{n}\tau_{i}$.  Let, $\tau_{-i}$ $=$ $\frac{\tau}{\tau_{i}}$.  Now suppose ($\overline{x}_{1}$, $\overline{x}_{2}$, $\ldots$, $\overline{x}_{n}$) is a solution to $\mathtt{MP2}$-$\mathtt{Dec}$.  Each $\overline{x}_{i}$ is a $\kappa$-step deterministic policy of the $i$th agent.  The $\kappa$-joint-policy formed by them is also deterministic.  If for a sequence $p$ of length $\kappa$,  $\overline{x}_{i}[p] = 1$,  then it implies that for exactly $\tau_{-i}$ joint-sequences $q$ of length $\kappa$ in which the sequence of the $i$th agent is $p$,  $\prod_{i = 1}^{n}\overline{x}_{i}[q] = 1$.  On the other hand, if $\overline{x}_{i}[p] = 0$,  then for each joint-sequence $q$ in which the sequence of the $i$ agent is $p$, $\prod_{i = 1}^{n}\overline{x}_{i}[q] = 0$.  This can be represented mathematically as, 
\begin{eqnarray}
\label{EquationJPC1}
\sum_{q \in {\cal{S}}^{\kappa} : q_{i} = p}\prod_{j = 1}^{n}\overline{x}_{j}[q_{j}] = \tau_{-i}\overline{x}_{i}[p], \mbox{$\quad$ $\forall$ $i \in N$, $\forall$ $p \in {\cal{S}}_{i}^{\kappa}$}
\end{eqnarray}
The set of equations (\ref{EquationJPC1})is true for every $\kappa$-step deterministic joint-policy,  and it allows us to linearize the constraints (\ref{EquationXToY}).  All we have to do is to add the following set of linear constraints to $\mathtt{MP2}$-$\mathtt{Dec}$,
\begin{eqnarray}
\label{EquationJPC2}
\sum_{q \in {\cal{S}}^{\kappa} : q_{i} = p}\overline{y}[q] = \tau_{-i}\overline{x}_{i}[p], \mbox{$\quad$ $\forall$ $i \in N$, $\forall$ $p \in {\cal{S}}_{i}^{\kappa}$}
\end{eqnarray}  
If these constraints are added to $\mathtt{MP2}$-$\mathtt{Dec}$ then the following holds,
\begin{eqnarray}
\prod_{j = 1}^{n}\overline{x}_{j}[q_{j}] = \overline{y}[q], \mbox{$\quad$ $\forall$ $q \in {\cal{S}}^{\kappa}$}
\end{eqnarray}  
because the r.h.s. of their corresponding equations are equal.  Thus, we have achieved the linearization of the constraints (\ref{EquationXToY}) and therefore of $f(\overline{x})$.  We shall call the constraints (\ref{EquationJPC2}) as the \textit{joint-policy constraints}.  The MP obtained on adding the joint-policy constraints to  $\mathtt{MP2}$-$\mathtt{Dec}$ gives us the integer linear program $\mathtt{ILP}$-$\mathtt{Dec}$,  on which the mixed ILP (MILP),  the main contribution of this paper,  is based  We give $\mathtt{ILP}$-$\mathtt{Dec}$ below for the sake of completeness.
\subsection{Integer linear program $\mathtt{ILP}$-$\mathtt{Dec}$}
\begin{enumerate}
\item Variables:
\begin{enumerate}
\item  A $|{\cal{S}}^{\kappa}|$-vector of variables, $\overline{y}$.  
\item  For each agent $i \in N$,  an $|{\cal{S}}_{i}|$-vector of variables, $\overline{x}_{i}$.
\end{enumerate}
\item  Objective:
\begin{eqnarray}
\mbox{maximize}&& f(\overline{y}) \equiv \sum_{q \in {\cal{S}}^{\kappa}}\nu(q)\overline{y}[q]
\end{eqnarray}
\item  Constraints:  for each agent $i \in N$,
\begin{enumerate}
\item  Policy constraints:   
\begin{equation}
\sum_{a_{i} \in A_{i}}\overline{x}_{i}[a_{i}] = 1
\end{equation}
$\forall$ $t \in \{1, 2, \ldots, \kappa - 1\}$, $\forall$ $p \in {\cal{S}}_{i}^{t}$, $\forall$ $o_{i} \in \Omega_{i}$,
\begin{equation}
\overline{x}_{i}[p] - \sum_{a \in A_{i}}\overline{x}_{i}[poa] = 0
\end{equation}
\item  Joint-policy constraints:  for each $p \in {\cal{S}}_{i}^{\kappa}$,
\begin{equation}
\sum_{q \in {\cal{S}}^{k} : q_{i} = p}\overline{y}[q] = \tau_{-i}\overline{x}_{i}[p]
\end{equation}
\end{enumerate}
\item  Integer constraints:
\begin{eqnarray}
\overline{x}_{i}[p] \in \{0, 1\}, &&\mbox{$\forall$ $i \in N$, $\forall$ $p \in {\cal{S}}_{i}$}\\
\overline{y}[q] \in \{0, 1\}, &&\mbox{$\forall$ $q \in {\cal{S}}^{\kappa}$}
\end{eqnarray}
\end{enumerate}
We thus have the following result.  
\begin{theorem}
An optimal solution ($\overline{x}_{1}$, $\overline{x}_{2}$, $\ldots$, $\overline{x}_{n}$) to $\mathtt{ILP}$-$\mathtt{Dec}$ yields an optimal $\kappa$-joint-policy for the given Dec-Pomdp.  {\rm{(Proof is omitted)}}
\end{theorem}

\subsection{Mixed integer linear program $\mathtt{MILP}$-$\mathtt{Dec}$}
An ILP is so called because it is an LP whose variables are constrained to take integer values.  In $\mathtt{ILP}$-$\mathtt{Dec}$,  each variable can be either $0$ or $1$.  The principle method for solving an integer linear program is \textit{branch and bound}.  So when solving $\mathtt{ILP}$-$\mathtt{Dec}$,  a tree of LPs is solved in which each LP is identical to the $\mathtt{ILP}$-$\mathtt{Dec}$ but in which the integer constraints are replaced by non-negativity constraints (i.e.,  all the variables are allowed to take real values greater than or equal to $0$).  In general,  the lesser the number of integer variables in an LP,  the faster a solution will be obtained.    Therefore it is desirable to minimize the number of integer variables in an LP.  An LP in which some variables are allowed to take real values while the remaining ones are constrained to be integers is called a mixed ILP (MILP).  Thus, an MILP may be solved faster than an ILP of the same size.  We say that an MILP is equivalent to an ILP if every solution to the MILP is also a solution to the ILP.  An MILP that is equivalent to $\mathtt{ILP}$-$\mathtt{Dec}$ can be conceived as follows.  Let this MILP be denoted by $\mathtt{MILP}$-$\mathtt{Dec}$.  Let $\mathtt{MILP}$-$\mathtt{Dec}$ be identical to $\mathtt{ILP}$-$\mathtt{Dec}$ in all respects except the following:  in each vector $\overline{x}_{i}$,  only those variables representing sequences of length $\kappa$ be constrained to take integer values $0$ or $1$;  \textit{all} the other variables in each $\overline{x}_{i}$ and \textit{all} the variables in the vector $\overline{y}$ be allowed to take real values greater than or equal to $0$.  Due to the equivalence,  we have the following result.
\begin{theorem}
An optimal solution ($\overline{x}_{1}$, $\overline{x}_{2}$, $\ldots$, $\overline{x}_{n}$) to $\mathtt{MILP}$-$\mathtt{Dec}$ yields an optimal $\kappa$-joint-policy for the given Dec-Pomdp
\end{theorem}
The proof of this theorem (and of the claim that $\mathtt{MILP}$-$\mathtt{Dec}$ is equivalent to $\mathtt{ILP}$-$\mathtt{Dec}$) is omitted due to lack of space.  The discussion henceforth applies to $\mathtt{ILP}$-$\mathtt{Dec}$ as well.  

\section{Improving $\mathtt{MILP}$-$\mathtt{Dec}$}
We now discuss two heuristics for improving the space and time requirement of formulating and solving $\mathtt{MILP}$-$\mathtt{Dec}$.
\subsection{Identifying dominated sequences}
The number of variables required in the $\mathtt{MILP}$-$\mathtt{Dec}$ can be minimized by using variables for only those sequences of each agent that are not \textit{dominated}.  Dominated sequences need not be represented in the $\mathtt{MILP}$-$\mathtt{Dec}$ because there always exists an optimal $\kappa$-joint-policy in which none of the policies contains a dominated sequence.  We first define dominated sequences of length $\kappa$.  Given sequences $p$ and $p'$ of length $\kappa$ of the $i$th agent,  $p'$ shall be called a \textit{co-sequence} of $p$ if it is identical to $p$ except for its last action.  Let ${\cal{C}}(p)$ denote the set of co-sequences of $p$.  Then, $p$ is said to be dominated if there exists a probability distribution $\theta$ over ${\cal{C}}(p)$,  such that for every joint-sequence $q$ of length $\kappa$ in which the sequence of the $i$th agent is $p$,  the following is true:
\begin{equation}
\nu(q) \leq \sum_{p' \in {\cal{C}}(p)}\theta(p')\nu(q')
\end{equation}
in which $q'$ = ($q_{1}$, $\ldots$, $q_{i - 1}$, $p'$, $q_{i + 1}$, $\ldots$, $q_{n}$).  Dominated sequences of length $\kappa$ can be identified through \textit{iterated elimination}.  Identifying sequences of lengths less than $\kappa$ is easier.  A sequence $p$ of length $t$ is a \textit{descendant} of a sequence $p''$ of length $j < t$ if the first $j$ actions and $j$ - 1 observations in $p$ are identical to the $j$ actions and $j$ - 1 observations in $p''$.  A sequence $p''$ of length $j$ is dominated if every descendant of $p''$ is dominated.  So, for each agent,  we first identify dominated sequences of length $\kappa$,  and then working backwards, we identify dominated sequences of lengths less than $\kappa$.   Note that \textit{if} dominated sequences are not represented by variables in $\mathtt{MILP}$-$\mathtt{Dec}$,  then in each joint-policy constraint the $=$ sign must be replaced by the $\leq$ sign.  The MILP that results when dominated sequences of all the agents are not represented by variables in $\mathtt{MILP}$-$\mathtt{Dec}$ and the above modifications are made shall be denoted by $\mathtt{MILP}$-$\mathtt{Pr}$-$\mathtt{Dec}$.

\subsection{Adding bounds into $\mathtt{MILP}$-$\mathtt{Dec}$}
The MILP solver can be guided in its path selection in the tree of LP problems or made to terminate as early as possible by providing lower and/or upper bounds on the objective function.  In this paper,  we wish to illustrate the importance of integrating bounds in $\mathtt{MILP}$-$\mathtt{Dec}$,  and so we have used rather loose bounds.  Given ${\cal{V}}(t)$,  the value of an optimal $t$-joint-policy,  a lower bound on the value of the optimal $(t + 1)$-joint-policy is,
\begin{equation}
\ell = {\cal{V}}(t) + \max\limits_{a \in A}\min\limits_{s \in S}R^{a}_{s}
\end{equation}   
For an upper bound, the value $u$ of an optimal $\kappa$-step policy of the Pomdp corresponding to the Dec-Pomdp can be used.  This value can be determined by the linear program (\ref{EquationLP1})-(\ref{EquationLP3}) which also finds the optimal $\kappa$-step policy for the Pomdp.  Let ${\cal{S}}^{t}$ denote the set of joint-sequences of length $t$.  Let $qoa$ denote the joint-sequence obtained on appending the joint-observation $o$ and the joint-action $a$ to the joint-sequence $q$.   
\begin{eqnarray}
\label{EquationLP1}
&&\mbox{maximize$\quad$} u = \sum_{q \in {\cal{S}}^{\kappa}}y[q] \mbox{$\quad$ s.t.:}\\
&&\sum_{a \in A}y[a] = 1\\
&& y[q] - \sum_{a \in A}y[qoa] = 0, \mbox{$\forall$ $t < \kappa$, $q \in {\cal{S}}^{t}$, $o \in \Omega$}\\
\label{EquationLP3}
&&y \geq 0
\end{eqnarray}
A bound is added to $\mathtt{MILP}$-$\mathtt{Dec}$ by adding a constraint.  The constraint $f(\overline{y}) \geq \ell$ is added for adding the lower bound and the constraint $f(\overline{y}) \leq u$ is added for adding the upper bound.

\section{Experiments}
\label{experiments}
\begin{table}
\begin{center}
\begin{tabular}{|l||c|c|c||c|c|}
\hline
 Algorithm & \multicolumn{3}{|c||}{MABC} & \multicolumn{2}{|c|}{MA-tiger}\\
 \hline
$\kappa$   &  3 & 4 & 5 & 3 & 4\\ 
\hline
\hline
$\mathtt{MILP}$-$\mathtt{Dec}$ & 0.86 & 900 & $-$ & 3.7& $\cdot$\\
$\mathtt{MILP}$-$\mathtt{Dec}$($u$) & 1.03 & 907 & $-$& 3.5 & $\cdot$ \\
$\mathtt{MILP}$-$\mathtt{Dec}$($\ell$) & 0.93 & 900 & $-$& 4.9 & 72\\
$\mathtt{MILP}$-$\mathtt{Pr}$-$\mathtt{Dec}$ & 0.84 & 80 & $\cdot$& 6.4& $\cdot$\\
$\mathtt{MILP}$-$\mathtt{Pr}$-$\mathtt{Dec}$($u$) & 0.93 & 10.2 & 25 & 6.2 & $\cdot$ \\
$\mathtt{MILP}$-$\mathtt{Pr}$-$\mathtt{Dec}$($\ell$) & 0.84 & 120 & $\cdot$& 7.6& 175\\
\hline
\hline
DP  & 5 & $$10$^{3}$ &  & & \\
MAA$^{*}$ & $t_{3}$ & $t_{4}$ &  & $t_{3}$ & $t_{4}$\\
PBDP & 1.0 & 2.0 & $$10$^{5}$ & $ t_{3}$ & $ t_{4}$\\
\hline
\hline
DP-JESP &  &  &  & $$ 0 & 0.02\\
Approx-DP &  &  &  & 0.05 & 1.0\\
MBDP & 0.01 & 0.01 & 0.02 & 0.46 & 0.72\\
\hline
\end{tabular}
\caption{Comparison of the runtimes in seconds of Dec-Pomdp algorithms.   $t_{3}$ denotes several seconds and $t_{4}$ denotes several hours.   ``$\cdot$'' denotes a time-out of $30$ minutes,  ``-'' denotes insufficient memory and blank denotes that the application of the concerned algorithm to the concerned problem does not appear in the literature.}
\label{Results}
\end{center}
\end{table}  
We formulated the MABC problem and MA-tiger problem as MILPs, and solved it using the ILOG Cplex 10 solver on an Intel P4 machine with 3.40 gigahertz processor speed and 2.0 GB ram.  The runtime in seconds of $\mathtt{MILP}$-$\mathtt{Dec}$ and $\mathtt{MILP}$-$\mathtt{Pr}$-$\mathtt{Dec}$ for different values of $\kappa$ is shown in Table \ref{Results}.  In the first column,  a parenthesis, if present indicates which bound is used.  The runtime includes the time taken to identify dominated sequences and compute the bound (for e.g., solve the LP for the Pomdp), where applicable.  We have listed the runtime of existing exact and approximate dynamic programming Dec-Pomdp algorithms as reported in the literature.  The three exact algorithms are DP, MAA$^{*}$ and PBDP.  The approximate algorithms are DP-JESP \cite{2003Nair}, Approximate-DP and MBDP.  As far as dominated sequences are concerned,  the MABC problem had about 75$\%$ dominated sequences per agent for $\kappa = 5$, while MA-Tiger had \textit{no} dominated sequences for any horizon.  

\section{Discussion and future directions}
In this paper we have introduced a new exact algorithm that for solving finite-horizon Dec-Pomdps.  The results from Table \ref{Results} show a clear advantage of the MILP algorithms over existing exact algorithm for the longest horizons considered in each problem.  We now point out three directions in which this work can be extended.
\begin{enumerate}
\item[] \textbf{Approximate algorithm:}  Our approach could be a good candidate to construct an approximate algorithm.  For instance,  if $\mathtt{MILP}$-$\mathtt{Dec}$ or one of its variant is able to solve a problem optimally for horizon $\kappa$ very quickly,  then it can be used as a ratchet for solving approximately for longer horizons in divisions of $\kappa$ steps.  Our initial experiments with this simple method on the MABC and MA-Tiger problems indicate that it may be comparable in runtime and value of the joint-policy found with current approximate algorithms for solving long horizons ($50$,$100$).  This is particularly  useful when the Dec-Pomdp problem cycles back to the original state in a few steps.  In the MA-Tiger problem, for example,  upon the execution of the optimal $3$-step joint-policy, denoted by $\sigma^{3}$,  the process returns back to its initial belief state.  The value of $\sigma^{3}$ is $5.19$.  So we can perpetually execute $\sigma^{3}$ to get in $m$ steps, a total expected reward of ($5.19m/3$).  Now,  the value of $\sigma^{2}$,  the optimal $2$-step joint-policy is $-2$.  For controlling the MA-Tiger problem for $m$ steps,  we may either (a) execute $\sigma^{3}$ $m/3$ times or (b) $\sigma^{2}$ $m/2$ times.  The loss for doing (b) instead of (a) would be $2.73/m$ per step.  This can be made arbitrarily high by changing the reward function.  In other words,  finding $\sigma^{3}$ is much more important that finding $\sigma^{2}$.  We can arrange for a similar difference in quality between $\sigma^{4}$ and $\sigma^{3}$;  and $\mathtt{MILP}$-$\mathtt{Dec}$ is able to find $\sigma^{4}$ in $72$ secs while other algorithms take hours.  Thus,  the role of an exact, fast algorithm, such as ours, may prove crucial even for very small problems.
\item[] \textbf{Dynamic programming:} In formulating $\mathtt{MILP}$-$\mathtt{Dec}$ we are required to first generate the set ${\cal{S}}_{i}^{\kappa}$ for each agent $i$.  The size of this set is exponential in $\kappa$.  The generation of this set acts as the major bottleneck for formulating $\mathtt{MILP}$-$\mathtt{Dec}$ in memory.  However,  we can use dynamic programming to create each set ${\cal{S}}_{i}^{\kappa}$ incrementally in a backward fashion.  Such a procedure does not require the knowledge of $b_{0}$ and it is based on the same principle as the DP algorithm.  In brief,  the procedure is explained as follows.  For each nonzero $t \leq \kappa$,  we generate for each agent a set of sequences of length $t$ by doing a \textit{backup} of a previously generated set of sequences of length $t$ - 1 of the agent.  We then compute for each joint-sequence of length $t$, an $|S|$-vector containing the values of the joint-sequence when the initial belief state is one of the states in $S$.  We then \textit{prune}, for each agent, sequences of length $t$ that are dominated over belief space formed by the cross-product of $S$ and the set of joint-sequences of length $t$.  By starting out with the set ${\cal{S}}_{i}^{1}$ (which is in fact just the set $A_{i}$) for each agent $i$,  we can incrementally build the set ${\cal{S}}_{i}^{\kappa}$.  Note that a backup of the set ${\cal{S}}_{i}^{t}$  creates $|A_{i}||\Omega_{i}||{\cal{S}}_{i}^{t}|$ new sequences;  i.e., the growth is linear.  In contrast, the backing-up of a set of policies represents an exponential growth.  The merit of this procedure is that we may be able to compute an optimal joint-policy for a slightly longer horizon.  But more importantly,  due to the linear growth of sequences in each iteration,  it may be possible to solve for the infinite-horizon by iterating until some stability or convergence in the values of joint-sequences in realized. 
\item[] \textbf{Pompds:}  Finally,  the approach consisting of the use of the sequence-form and mathematical programming could be applied to Pomdps.  We have already shown in this paper how a finite-horizon Pomdp can be solved.  In conjunction with the dynamic programming approach analogous to the one described above, it may be possible to compute the infinite-horizon discounted value function of a Pomdp.
\end{enumerate}

\section*{Acknowledgements}
We are grateful to the anonymous reviewers for providing us with valuable comments on this work and suggestions for improving the paper.

\bibliographystyle{aaai}

\end{document}